\documentclass[a4paper, 10pt, conference]{ieeeconf}      

\IEEEoverridecommandlockouts                              

\usepackage{graphics} 
\usepackage{tabularx}
\usepackage{amsmath} 
\usepackage{amssymb}  

\usepackage{bm}

\title{\LARGE \bf
Stable In-hand Manipulation for a Lightweight Four-motor \\Prosthetic Hand
}

\author{Yuki Kuroda$^{1}$, Tomoya Takahashi$^{1}$, Cristian C. Beltran-Hernandez$^{1}$, \\ Kazutoshi Tanaka$^{1}$ and Masashi Hamaya$^{1}$
\thanks{$^{1}$OMRON SINIC X Corporation, Tokyo, Japan. 
        {\tt\small yuki.kuroda@sinicx.com}}%
}

\usepackage{tabularx}
\usepackage{multirow}
 \usepackage[table,xcdraw]{xcolor}
 \usepackage{gensymb}
\usepackage{amsmath}
\usepackage{mathcomp}
\usepackage{mathtools}
\usepackage{rotating}
\usepackage{float}
\usepackage{pdflscape}
\usepackage{amsmath}
\usepackage{amssymb}
\usepackage{amsfonts}
\usepackage{bm}

\usepackage{booktabs} 
\usepackage{siunitx}  
\usepackage{multirow} 
\usepackage{tabularx} 
\usepackage{makecell}

\usepackage{boldline}

\let\labelindent\relax
\usepackage{enumitem}

\usepackage{threeparttable}
\usepackage{tikz}

\begin{document}

\maketitle
\thispagestyle{empty}
\pagestyle{empty}

\begin{tikzpicture}[remember picture, overlay]
  \node[anchor=south, yshift=5mm] at (current page.south) {
    \parbox{\textwidth}{\centering \footnotesize
      \copyright 2026 IEEE. Personal use of this material is permitted.  Permission from IEEE must be obtained for all other uses, in any current or future media, including reprinting/republishing this material for advertising or promotional purposes, creating new collective works, for resale or redistribution to servers or lists, or reuse of any copyrighted component of this work in other works.
    }
  };
\end{tikzpicture}

\begin{abstract}

Electric prosthetic hands should be lightweight to decrease the burden on the user, shaped like human hands for cosmetic purposes, and designed with motors enclosed inside to protect them from damage and dirt.
Additionally, in-hand manipulation is necessary to perform daily activities such as transitioning between different postures, particularly through rotational movements, such as reorienting a pen into a writing posture after picking it up from a desk.
We previously developed PLEXUS hand (Precision--Lateral dEXteroUS manipulation hand), a lightweight (311~g) prosthetic hand driven by four motors. This prosthetic performed reorientation between precision and lateral grasps with various objects. However, its controller required predefined object widths and was limited to handling lightweight objects (of weight up to 34 g).
This study addresses these limitations by employing motor current feedback. Combined with the hand’s previously optimized single-axis thumb, this approach achieves more stable manipulation by estimating the object’s width and adjusting the index finger position to maintain stable object holding during the reorientation.
Experimental validation using primitive objects of various widths (5--30~mm) and shapes (cylinders and prisms) resulted in a 100\% success rate with lightweight objects and maintained a high success rate ($\geqq$80\%) even with heavy aluminum prisms (of weight up to 289 g). By contrast, the performance without index finger coordination dropped to just 40\% on the heaviest 289~g prism. The hand also successfully executed several daily tasks, including closing bottle caps and orienting a pen for writing.
\end{abstract}

\section{INTRODUCTION}
\label{sec:introduction}

\begin{figure}[t]
  \centering
  \includegraphics[width=\linewidth]{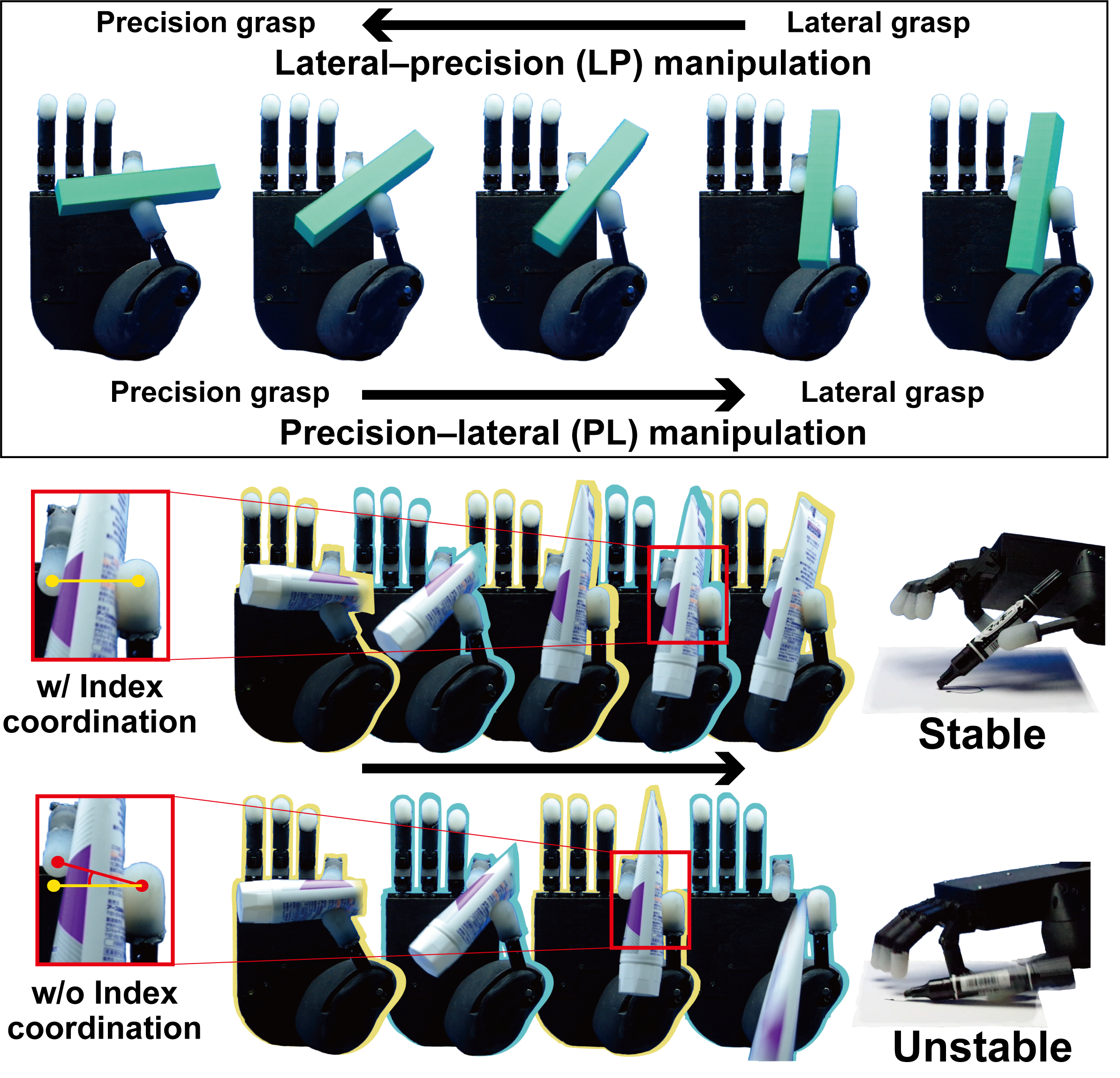}
  \caption{PL and LP in-hand manipulation. To effectively support the thumb's force, the index finger shifts its position using motor current feedback, as illustrated by the change from the red (without (w/o) the use of the index) to the yellow (with (w/) the use of the index) line in the red insets. This enables both stable object manipulation and a firm grip for tool use (w/ index), whereas a lack of coordination leads to grip failure (w/o index).}
  \label{fig:teaser}
  \vspace{3ex}
\end{figure}

Electric prosthetic hands restore functionality in individuals who have lost their hands owing to congenital anomalies or acquired causes. For user acceptance and practical daily use, key design considerations of these prosthetic hands include the following: a human-like, five-fingered appearance for cosmetic reasons~\cite{Carroll2004}, lightweight design (ideally less than 500 g~\cite{Vinet1995}) to reduce the physical burden on the user, internal motors for dirt and damage protection~\cite{Mohammadi2020}, and the ability to support diverse activities of daily living (ADL)~\cite{BelterDollar2011}. ADLs include fundamental self-care and household tasks involving grasping and in-hand manipulation. One particularly critical capability, which we term ``precision--lateral (PL) manipulation,’’ is the ability to switch between precision and lateral grasps smoothly without re-grasping (see Fig.~\ref{fig:teaser}). This function is crucial for prosthetic users, particularly for those with bilateral limb losses or an occupied unaffected hand, enabling single-handed task completion (e.g., closing a bottle cap and reorienting a pen for writing). Practical application demands stable manipulation of objects varying in shape and weight, and a final grasp sufficiently firm to resist external forces, such as during writing (as illustrated in Fig.~\ref{fig:teaser}).

Despite recent advancements, lightweight five-finger prosthetic hands (e.g., the bebionic hand~\cite{OttobockUS2023}: 14 grip patterns and 495--529 g, and the i-Limb~\cite{Ossur2023}: 36 grip patterns and 454--628 g) still primarily provide only static grasping. In contrast, in-hand manipulation typically relies on dexterous hands with many degrees of freedom (DOFs), such as the Shadow Dexterous Hand (20 motors)~\cite{OpenAI2019}, TWENDY-ONE hand (13 motors)~\cite{Funabashi2018}, and the Allegro Hand (16 motors)~\cite{Or2016}. In these hands, a high-DOF mechanism and dense sensory information form the basis for dexterous manipulation control. However, this hardware-intensive approach results in substantial weight ($>$1500~g) and complexity, rendering it impractical for the lightweight, self-contained design required by prosthetics. Consequently, a technology that can achieve dexterous in-hand manipulation while satisfying the demanding constraints of electric prostheses has not yet been established.

To address this challenge, we previously developed a lightweight prosthetic hand that enables stable PL in-hand manipulation. Our proposed PLEXUS hand (Precision--Lateral dEXteroUS manipulation hand)~\cite{KurodaICORR2025} is a 311 g four-motor prosthetic hand featuring a single-axis thumb that rotates around a carpometacarpal (CM) joint.
We computationally optimized the position of the CM joint so that the PLEXUS hand could perform PL manipulation of various objects.
However, in a previous study~\cite{KurodaICORR2025}, we were required to input a predefined object width for the open-loop controller. Additionally, the study dealt with relatively lightweight objects (up to 34.34 g).

This study addressed the aforementioned issue by employing motor current feedback, which works in conjunction with the hand’s previously optimized single-axis thumb for more stable and practical PL manipulation.
Because each finger is actuated by a single motor, the motor current provides a clear indication of the contact state, enabling threshold-based detection without the need for additional sensors.
Our current feedback has two roles. First, it is used for object width estimation during the precision grasp; second, based on the estimated width, it strategically coordinates the index finger to apply a sufficient force, a method inspired by human hand mechanics used for applying torques~\cite{Shim2007}. This feature enables the stable PL manipulation of heavy objects without the manual inputs of the object width, addressing limitations of previous methods~\cite{KurodaICORR2025}.

We validated the performance and force application capabilities of this hand via PL manipulation evaluations that included heavy items. The hand achieved a 100\% success rate with lightweight objects (cylinders and prisms, 5--30 mm) and maintained a high success rate ($\geqq$80\%) even with heavy aluminum prisms (up to 289 g). Furthermore, we demonstrated the practical utility of the hand in several ADL tasks, including closing bottle caps and orienting a pen for writing, highlighting its potential for daily use.

The main contributions of this study are as follows: first, an estimation method for object widths is introduced using only motor current feedback. The feedback is based on a simplified mechanism for PL in-hand manipulation (which eliminates the need for manual input required by previous open-loop systems). Second, an index finger coordination strategy is proposed based on a human-inspired approach that uses this estimated width to adapt the grasp and apply sufficient force, enabling the stable manipulation of heavy objects.

\section{Introduction of PLEXUS Hand}
The core design objective of the PLEXUS hand is to enable both fundamental grasping~\cite{Cipriani2008, Cipriani2010} and PL manipulation based on a minimal number of actuators using an optimized single-axis thumb. The design of the four fingers (index, middle, ring, and little fingers), including their basic configuration and actuation mechanisms (e.g., four-bar linkages), was based on approaches detailed in previous studies, such as that of Wang et al.~\cite{Wang2022}, and these finger configurations and actuation mechanisms were considered as fixed parameters in our study. Additional details regarding the optimization and the overall mechanical design of the PLEXUS hand, including the finger mechanisms, can be found in our previous publication~\cite{KurodaICORR2025}. This section introduces the PLEXUS hand and its open-loop control method.

\begin{figure}[t]
\center{
\includegraphics[keepaspectratio, width=1.0\linewidth]{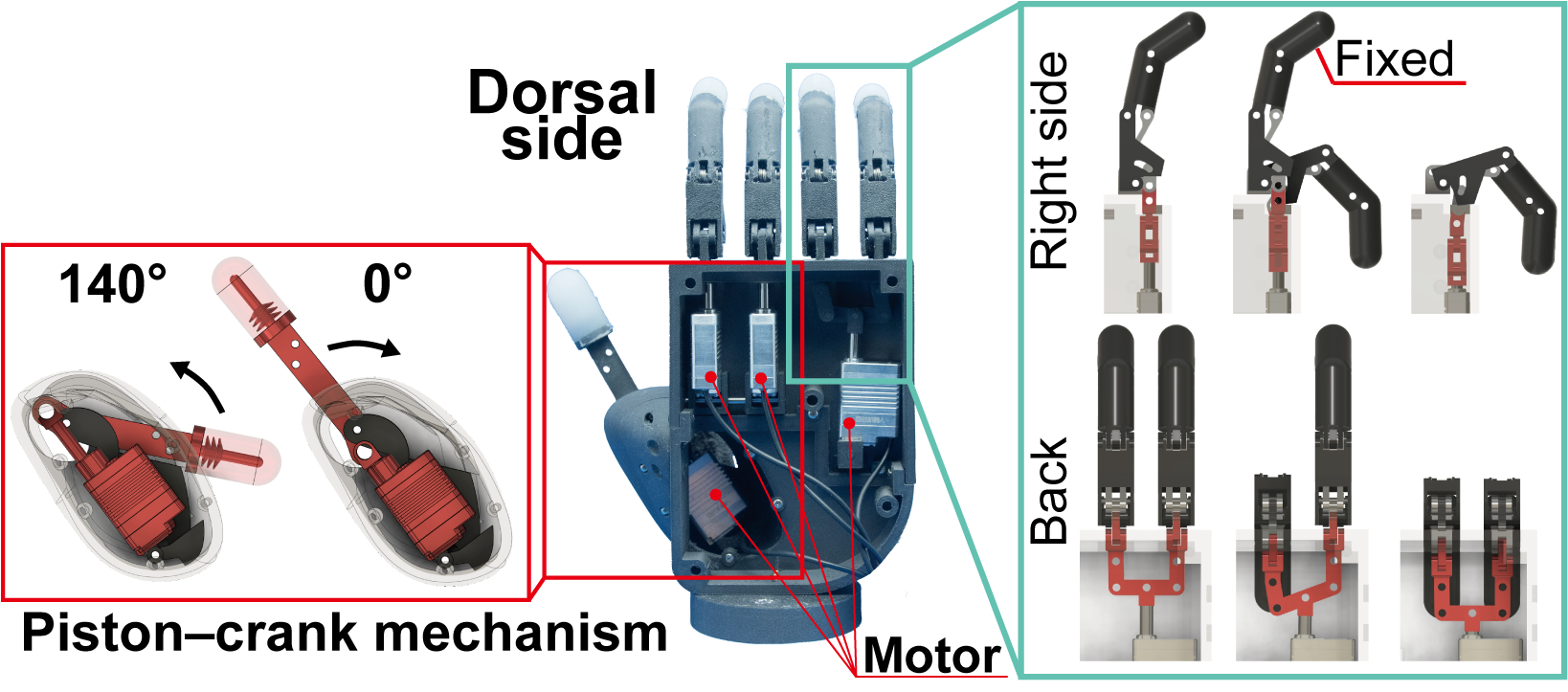} 
\caption{Actuation mechanisms for the thumb, ring, and little fingers~\cite{KurodaICORR2025}.
In the case of the thumb, rotation is driven by a linear actuator operated based on a piston--crank mechanism. For the ring and little fingers, a single actuator drives both digits via a differential Y-shaped linkage.}
\label{fig:GraspRequirements}
}
\vspace{3ex}
\end{figure}

\subsection{PLEXUS Hand}
We engineered and fabricated a PLEXUS hand prototype based on the optimization process described in our previous publication~\cite{KurodaICORR2025}. This hardware implementation embodies the optimal thumb-axis position determined by the proposed algorithms. As demonstrated in our previous study~\cite{KurodaICORR2025}, the PLEXUS hand is capable of performing five basic grasp postures essential for ADLs in addition to the PL manipulation capability, which is the focus of the current study (shown in Fig.~\ref{fig:teaser}). 

The PLEXUS hand employs distinct actuation mechanisms for different digit groups, all of which are driven by linear actuators (LAS16-023D and LAS10-023D, Beijing Inspire Robots Technology, Beijing, China) for a simplified system integration. The specific designs of these mechanisms, including the piston--crank mechanism for the thumb, the four-bar linkage of the index and middle fingers, and the differential mechanism for the ring and little fingers (Fig.~\ref{fig:GraspRequirements}), are described in our previous study~\cite{KurodaICORR2025}.

\subsection{Open-loop Control Based on a Predefined Object Width}
In our preliminary work~\cite{KurodaICORR2025}, the PLEXUS hand was operated by an open-loop controller that relied on a predefined object width $w$. The manipulation sequence was as follows:

\begin{enumerate}
\item The object width $w$ is input to the controller before initiating the grasp.
\item When the user commands an initial precision grasp, the controller drives the thumb and index finger. The thumb stops at a predefined angle $\theta_\mathrm{T,P}$ directly below the index finger.
\item The index finger stops at a target angle $\theta_{\text{c}}$, determined by a precalibrated mapping function $f': w \mapsto \theta_{\text{c}}$. This mapping $f'$ is derived beforehand based on the optimization algorithm described in~\cite{KurodaICORR2025}.
\item The controller then uses the input width $w$ and a second predefined mapping $g': w \mapsto \theta_\mathrm{T,L}$ (also based on the optimization algorithm) to determine the target joint angle $\theta_\mathrm{T,L}$ for the thumb in the target lateral grasp posture.
\end{enumerate}

This open-loop approach, which lacks feedback on contact or force states, struggled to apply appropriate force to prevent object ejection and to coordinate the fingers to optimal positions for a stable hold. Consequently, the hand's manipulation capabilities were limited.

\section{PL Manipulation Using Motor Current Feedback}\label{sec:Control}
When using the PLEXUS hand developed in this study as a myoelectric prosthesis, the user is assumed to control the hand via surface electromyogram (EMG) sensors. One plausible control method is an EMG pattern recognition system~\cite{KurodaCBS2025} in which the user learns to generate specific muscle contraction patterns corresponding to several target grasp postures (e.g., rest, precision grasp, lateral grasp, and power grasp), and selects the desired posture by intentionally producing the learned pattern. However, as suggested in~\cite{KurodaCBS2025}, the number of patterns a user can reliably discriminate and control in practical situations is limited and is reported to be approximately four.

Even if the user can select a target grasp posture, directly controlling the complex coordinated finger movements required for the transition from one posture to another while holding an object (i.e., in-hand manipulation) in real-time via EMG signals alone is extremely difficult.

The system implemented in this study controls the transition process between the grasp postures. Its implementation and the relative simplicity of its control logic, which avoids relying on sophisticated sensor systems (e.g., dense tactile arrays and vision) or machine-learning algorithms often used for controlling dexterous manipulation~\cite{OpenAI2019, Funabashi2018, Handa2023, Yin2023}, are characteristic features. This is made possible because the core hardware design of the hand---the optimized single-axis thumb---simplifies the control problem by mechanically generating the entire motion path required for PL manipulation. To complement this mechanical design, the control logic incorporates a strategic, human-inspired coordination of the index finger~\cite{Shim2007} that provides the stability needed to manipulate heavy objects. This current-based control system was developed to overcome the limitations of our preliminary work~\cite{KurodaICORR2025}, which relied on an open-loop scheme and could not provide sufficient force for manipulating heavy objects.

\begin{figure}[t]
    \centering
    \includegraphics[width=1.0\linewidth]{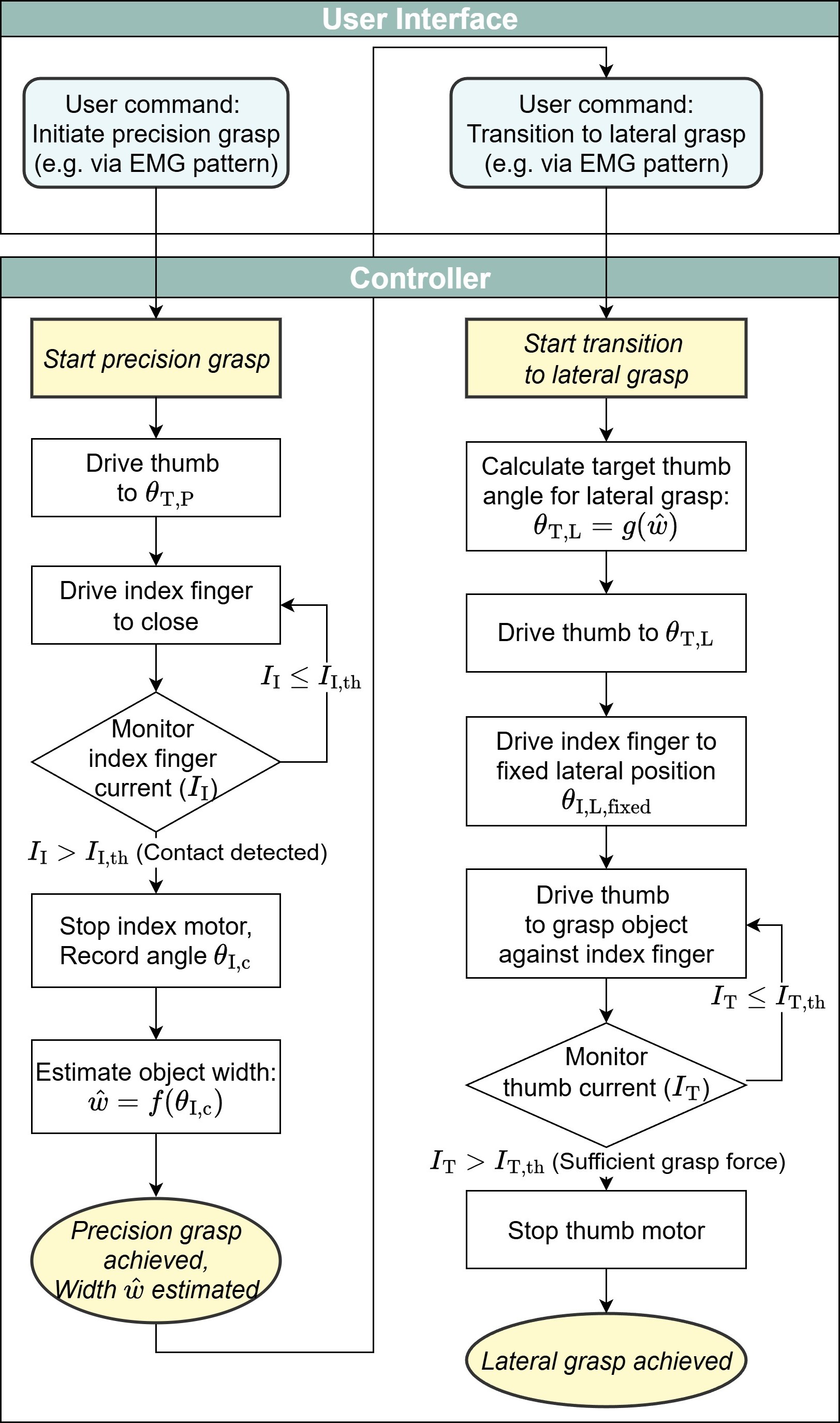}
    \caption{Schematic of the control system for the PL manipulation of the PLEXUS Hand. The flowchart illustrates the logic for estimating the object width using the index finger motor current and subsequently coordinating the thumb and index finger movements to facilitate object reorientation.}
    \label{fig:ControlFlowProcess} 
\end{figure}

This autonomous transition system is activated when the user issues a high-level command, such as selecting the EMG pattern corresponding to the intent to transition from a precision grasp to a lateral grasp. The system then executes the following control flow using only the motor-current monitoring function (which is commonly available on the actuators) as the sensory input. The control flow primarily comprises two stages: an initial grasp based on an estimation of the object width, and subsequent width-adaptive finger coordination (see the schematic in Fig.~\ref{fig:ControlFlowProcess}). To clarify the variables used in this control system, we adopt a unified subscript notation. Subscripts, written in Roman type (e.g., $\theta_{\mathrm{T,P}}$), denote the relevant finger and the context. The primary symbols are defined as: $\mathrm{T}$ (Thumb), $\mathrm{I}$ (Index finger), $\mathrm{P}$ (Precision grasp), $\mathrm{L}$ (Lateral grasp), $\mathrm{c}$ (contact), and $\mathrm{th}$ (threshold). Thus, $\theta_{\mathrm{T,P}}$ denotes the Thumb's angle for the Precision grasp, and $I_{\mathrm{I,th}}$ denotes the current threshold for the Index finger.

\subsection{Stage 1: Initial Grasp and Object Width Estimation}
\begin{enumerate}
    \item When the user commands an initial precision grasp, the controller drives the thumb and index finger in the closing direction. The thumb stops at a predefined angle $\theta_\mathrm{T,P}$ directly below the index finger.
    \item As the fingers contact the object and motor load increases, the motor current ($I_{\text{I}}$) of the index finger increases abruptly due to the low backdrivability of the actuators used in this hand.
    \item The controller monitors $I_{\text{I}}$ and detects contact when it exceeds a predefined threshold ($I_{\mathrm{I,th}}$ = 300 mA).
    \item Upon contact detection, the controller stops the driving motor of the index finger and records the corresponding joint angle $\theta_{\mathrm{I,c}}$. Notably, a low actuator backdrivability indicates that current spikes are useful only for detecting the initial object contact (not for monitoring the static hold), as the mechanism maintains the position without a sustained high current.
    \item An estimated object width $\hat{w}$ is determined from the recorded angle using a precalibrated mapping function $f: \theta_{\mathrm{I,c}} \mapsto \hat{w}$. This function $f$ maps the direct kinematic relationship between the index finger's contact angle ($\theta_{\mathrm{I,c}}$) and the object width ($\hat{w}$), which is geometrically defined when the thumb is positioned at its pre-grasp position $\theta_\mathrm{T,P}$. In practice, this mapping is implemented as a discrete lookup table, interpolating between pre-calibrated angle-to-width values, corrected for offsets measured on the actual hardware.
\end{enumerate}

\subsection{Stage 2: Width-adaptive Finger Coordination}
This stage implements a human-inspired coordination strategy~\cite{Shim2007}. The index finger is actively positioned to create a stable pivot, which is critical for the manipulation of heavy objects. The control sequence is as follows:
\begin{enumerate}
    \item When the user commands the transition to the next target posture (i.e., lateral grasp), the autonomous manipulation sequence begins.
    \item The controller uses the estimated width $\hat{w}$ and a predefined mapping $g: \hat{w} \mapsto \theta_\mathrm{T,L}$ to determine the target joint angle $\theta_\mathrm{T,L}$ for the thumb in the target lateral grasp posture. This mapping $g$ is derived from the optimization process in~\cite{KurodaICORR2025}, which pre-calculates the optimal angle $\theta_\mathrm{T,L}$ that satisfies the lateral grasp stability constraints (as described in~\cite{KurodaICORR2025}) for a given object width $\hat{w}$. As with $f$, this function is implemented as a lookup table storing the pre-computed angles for discrete object widths (e.g., at 5~mm width intervals).
    \item The controller first drives the thumb from its current position to the calculated target angle $\theta_\mathrm{T,L}$.
    \item Inspired by human strategies~\cite{Shim2007}, the controller drives the index finger to the stationary target angle $\theta_\mathrm{I,L,fixed}$, which is determined by empirically refining a theoretically optimal posture. The optimization process in~\cite{KurodaICORR2025} first provided this theoretical ideal for a firm grasp, which corresponds to the intersection of the thumb line of action and the index-finger center. We then fine-tuned this posture via experimental trials to identify the final, robust angle used for the manipulation tasks.
    \item After the index finger reaches its target position, the controller drives the thumb again to grasp the object firmly against the index finger. During this final grasp closure, the thumb motor current is monitored, and the drive stops when it exceeds a predefined grasping threshold ($I_{\mathrm{T,th}}$ = 400 mA), corresponding to the desired holding force.
\end{enumerate}

The transition from a lateral to a precision grasp follows an analogous procedure, but with reversed finger roles. Notably, the final current-based grasping closure (step 5) is omitted for lateral--precision (LP) manipulation to prevent potential object ejection, given the kinematics of the precision grasp posture, where excessive force could misalign the fingertips.

\section{Experiments}
We conducted two experiments to evaluate the performance and practical applicability of the PLEXUS hand equipped with the proposed current-based control system, as described in Section~\ref{sec:Control}. These tasks were required to be executed autonomously by the hand’s controller in response to a high-level user command.

Experiment 1 aimed to assess quantitatively two key advancements in the PL manipulation capabilities of the PLEXUS hand. The first objective was to evaluate the performance of the system in adapting to objects of varying widths without the manually preset width information required in our previous study. The hand was required to reorient objects with flat or curved surfaces up to 30~mm in width while maintaining a stable hold without dropping, a target selected to exceed the median width of objects typically handled in precision grasps during daily activities~\cite{Feix2014}. The second objective was to determine whether a relatively simple control strategy, namely the coordination of the index finger based on this width estimation, could overcome the grasp stability and forceful manipulation limitations previously encountered with a stationary index finger. The hand was required to manipulate objects weighing up to 150~g, a target selected to exceed the median weight of objects handled in precision grasps in daily life~\cite{Feix2014}.

Experiment 2 demonstrated the utility of the hand in several representative daily tasks that require object reorientation. Specifically, the experiment was designed to confirm two points: whether the index finger coordination strategy is effective in daily tasks, and whether these tasks can be successfully performed with a high-level user command, such as activating a switch for a lateral grasp.

\subsection{Experiment 1: Quantitative Evaluation of PL Manipulation}

\subsubsection{Evaluation Tasks}
The experiment focused on the execution of PL manipulation in both directions. Three control conditions were evaluated.
\begin{itemize}
    \item \textbf{PL Manipulation with Index Finger Coordination (PL w/ index):} Transition from precision to lateral grasp using the full control strategy, including index finger movement.
    \item \textbf{PL Manipulation without Index Finger Coordination (PL w/o index):} Transition from precision to lateral grasp during which the index finger remained stationary at its initial contact angle. This condition isolates the effect of the index finger coordination by comparing its results to PL w/ index.
    \item \textbf{LP Manipulation with Index Finger Coordination (LP w/ index):} Transition from lateral to precision grasp. As the index finger coordination is essential for this direction, only this condition was evaluated for LP manipulation to assess its absolute performance.
\end{itemize}

\subsubsection{Evaluated Objects and Materials}
The objects evaluated included primitive shapes and common daily items.
\begin{itemize}
    \item \textbf{Primitive Objects:} Cylinders and square prisms (height: 120~mm) with varying widths and diameters (5, 10, 15, 20, 25, 30~mm) were evaluated. Both lightweight polylactic acid versions and heavier aluminum versions were used to assess the performance under different inertial loads and surface conditions.
    \item \textbf{Common Objects:} A diverse set of items representative of ADLs were evaluated, including tools (e.g., precision screwdriver, seal stamp), stationery items (e.g., pen, eraser), and household items (e.g., card, spoon).
\end{itemize}

\subsubsection{Procedure}
Ten trials were conducted for each object for each relevant task condition (PL w/ index, PL w/o index, LP w/ index). This sample size was chosen to demonstrate the feasibility of the proposed method and is consistent with evaluation methodologies employed in related previous studies on PL manipulation~\cite{Or2016, Funabashi2018}. The procedure for each trial was as follows:
\begin{enumerate}[label={S\arabic*})]
    \item The trial started with the hand in an open state.
    \item An initial grasp (precision for PL w/ index and PL w/o index tasks; lateral for LP w/ index task) was commanded. The system performed Stage 1: grasping the object via motor current feedback and estimating its width $\hat{w}$.
    \item \textit{Initial Object Placement and Stability Check:} During the initial grasp (S2), the experimenter visually adjusted the object placement configuration according to the task type. For the initial precision grasp (PL tasks), the object was adjusted to ensure that it was (i) grasped near its center of gravity, (ii) oriented perpendicular to the thumb rotation axis, and (iii) the object center was positioned between the relevant fingertip centers. For the initial lateral grasp (LP task), the object was adjusted to ensure that it was (i) grasped near its center of gravity, (ii) oriented parallel to the central axis of the index fingertip, and (iii) positioned between the side surfaces of the thumb and index finger. After the grasping process was completed, the experimenter lightly perturbed the object by hand to confirm the grasp stability before proceeding.
    \item Subsequently, a command to transition to the target posture (lateral for PL tasks, precision for an LP task) was issued. The system performed Stage 2, executing finger coordination according to the specified task condition as described in Section~\ref{sec:Control}. Success was evaluated at this point based on the criteria described below.
    \item Finally, the grasp was released, and the hand returned to the open state.
\end{enumerate}
The sequence from the initial grasp command in S2 through the completion of the transition in S4 (excluding the experimenter check in S3) was executed autonomously in response to the respective commands.

\subsubsection{Success Criteria}
A trial was considered successful if after completing the transition movement in S4 and reaching the target posture, the object was stably held (did not drop or slip), as confirmed via visual inspection by the experimenter. Even if the position of the object shifted during manipulation owing to factors such as weight, the trial was still considered successful if the object was ultimately stably grasped (and remained within the hand, that is, it did not drop).

\begin{table}[t!]
\centering
\begin{threeparttable} 
\caption{Success rates (\%) for manipulation tasks over 10 trials.} 
\label{tab:Exp1_Results_Combined}
\begin{tabular}{@{} p{1.8cm} p{0.7cm} p{0.7cm} S[table-format=3.2, table-column-width=0.7cm] c c c @{}}
\toprule
Object type & Width & Material & {Weight} & \multicolumn{3}{c}{Success rate (\%)} \\
\cmidrule(lr){5-7}
 & {(mm)}&& {(g)}& \multicolumn{1}{p{0.7cm}}{\centering PL w/ index} & \multicolumn{1}{p{0.7cm}}{\centering PL w/o index} & \multicolumn{1}{p{0.7cm}}{\centering LP w/ index} \\
\midrule
\multicolumn{7}{l}{\textit{Primitive objects}} \\ \midrule
Cylinder           & 5               & PLA      & 2.79   & 100 & 100 & 100 \\
Square prism       & 5               & PLA      & 2.49   & 100 & 100 & 100 \\
Cylinder           & 10              & PLA      & 5.04   & 100 & 100 & 100 \\
Square prism       & 10              & PLA      & 6.20   & 100 & 100 & 100 \\
Cylinder           & 15              & PLA      & 9.90   & 100 & 90  & 100 \\
Square prism       & 15              & PLA      & 11.59  & 100 & 100 & 100 \\
Cylinder           & 20              & PLA      & 14.58  & \textbf{100} & 40  & 80  \\
Square prism       & 20              & PLA      & 18.03  & 100 & 100 & 100 \\
Cylinder           & 25              & PLA      & 20.80  & 100 & 80  & 80  \\
Square prism       & 25              & PLA      & 24.65  & 100 & 100 & 100 \\
Cylinder           & 30              & PLA      & 28.34  & 100 & 100 & 80  \\
Square prism       & 30              & PLA      & 34.34  & 100 & 100 & 100 \\ \midrule
Cylinder           & 5               & Al & 6.25   & 100 & 100 & 90  \\
Square prism       & 5               & Al & 8.04   & 100 & 100 & 100 \\
Cylinder           & 10              & Al & 24.92  & 100 & 100 & 100 \\
Square prism       & 10              & Al & 32.18  & 100 & 80  & 100 \\
Cylinder           & 15              & Al & 56.48  & \textbf{80}  & 40  & 40  \\
Square prism       & 15              & Al & 72.18  & \textbf{90}  & 60  & 100 \\
Cylinder           & 20              & Al & 100.31 & 70  & 70  & 30  \\
Square prism       & 20              & Al & 128.53 & 100 & 100 & 100 \\
Cylinder           & 25              & Al & 158.41 & \textbf{50}  & 0   & 10  \\
Square prism       & 25              & Al & 200.66 & 100 & 90  & 100 \\
Cylinder           & 30              & Al & 228.00 & 40  & 20  & 10  \\
Square prism       & 30              & Al & 289.09 & \textbf{80}  & 40  & 100 \\ \midrule
\multicolumn{7}{l}{\textit{Common Objects}} \\ \midrule
Business card               & 0.3             & {---}    & 1.16   & 0   & 20  & 100 \\
Heavy pen          & 10.3            & {---}    & 23.45  & 100 & 100 & 100 \\
Light pen          & 10.2            & {---}   & 9.75   & 100 & 100 & 100 \\
Seal stamp         & 9.9             & {---}   & 4.68   & 100 & 100 & 100 \\
Spoon              & 1.7             & {---}    & 15.48  & 20  & 0   & 100 \\
Screwdriver (\#1)  & 6.2             & {---}    & 21.22  & 100 & 100 & 100 \\
Knife              & 10.5--12.5            & {---}    & 63.03  & \textbf{100} & 0   & 100 \\
Thick spoon        & 10.15           & {---}    & 48.08  & 100 & 100 & 100 \\
Thick teaspoon    & 8.55            & {---}    & 26.63  & 100 & 100 & 100 \\
Thick fork         & 10.15           & {---}    & 32.44  & 100 & 100 & 100 \\
Eraser             & 12.6            & {---}   & 39.11  & 100 & 90  & 100 \\
Dishwashing det.   & 43.86           & {---}  & 207.34 & \textbf{50}  & 0   & 80  \\
Toothpaste         & 20--24           & {---}  & 98.22  & \textbf{100} & 0   & 100 \\
Bolt (M12-80~mm)    & 11.81           & {---}    & 75.54  & 50  & 50  & 30  \\ \bottomrule
\end{tabular}
\begin{tablenotes}
    \footnotesize
        \item[*] PL: Precision--lateral; LP: Lateral--precision; w/: with; w/o: without; Al: Aluminum; PLA: Polylactic acid.
\end{tablenotes}
\end{threeparttable}
\end{table}

\subsection{Experiment 2: Demonstration of Practical Applications}
Although comprehensive testing involving prosthesis users remains essential for future work, we conducted preliminary demonstrations in several practical scenarios to demonstrate qualitatively the potential utility of the manipulation capability of the PLEXUS hand (specifically, using the PL w/ index strategy) in tasks representative of ADLs that benefit from object reorientation. 

\begin{figure}[t!]
\centering
\includegraphics[width=\linewidth]{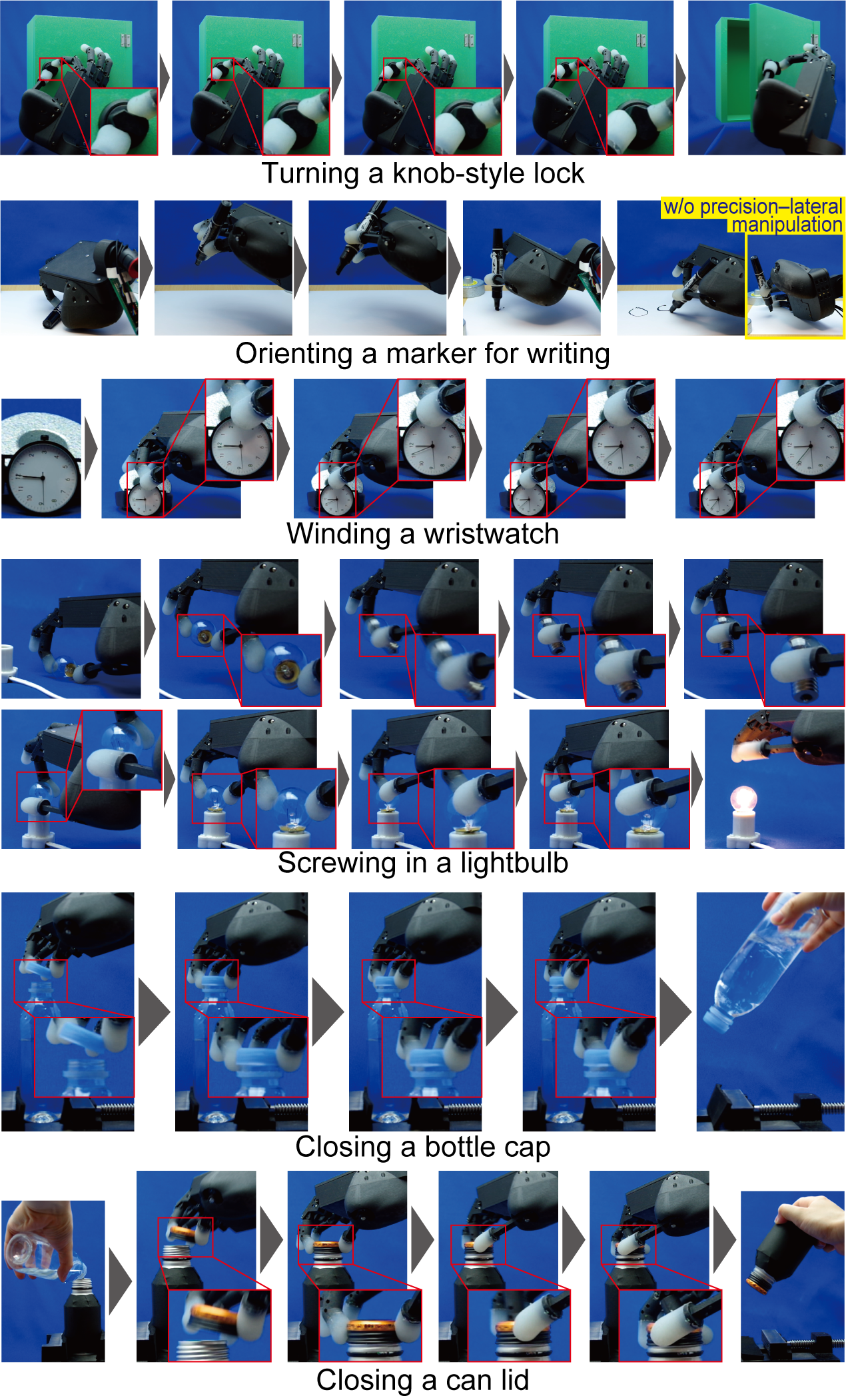}
\caption{Demonstrations of dexterous daily tasks with the PLEXUS Hand. The figure illustrates PL manipulation (PL w/ index strategy) for tasks such as knob turning and bottle cap closing. The yellow frame highlights an instance where manipulation is essential to avoid palmar obstruction. See supplementary Movie S1 (https://youtu.be/TswoBs84uyE) for full sequences.}
\label{fig:Demonstrations}
\end{figure}

These demonstrations involved an operator supporting the base of the hand using a selfie stick to position it for the task. This setup is intended to simulate the actual usage environment of a prosthesis, where the user must actively adjust the hand's position for tasks. Therefore, the goal of this experiment was not to measure quantitative success but to qualitatively confirm the feasibility of the manipulation itself in representative ADLs. Success was assessed qualitatively based on task completion. These included dexterous daily tasks requiring PL manipulation, including knob turning and bottle cap closing, as shown in Fig.~\ref{fig:Demonstrations}. 

\section{Results and Discussion}

\subsection{Experiment 1: Quantitative Evaluation of PL Manipulation}

The success rates for the manipulation tasks using primitive and common objects are summarized in Table~\ref{tab:Exp1_Results_Combined}, where boldface fonts are used to highlight cases in which the two PL conditions differ by 30 percentage points or more. The quantitative results from Experiment 1 demonstrate the effectiveness of the proposed hand while also highlighting the benefits and limitations of the approach.

The high success rates obtained using the primary control strategy (PL w/ index) were observed for most lightweight primitive objects (PLA cylinders and prisms), indicating that the combination of the optimized single-axis thumb design and relatively simple current-based control is sufficient for achieving reliable PL manipulation in basic scenarios. This finding demonstrates that a meticulous mechanical design can simplify the control requirements for specific manipulation tasks. This contrasts with strategies that rely solely on complex sensing and control architectures typical of high-DOF hands (discussed in Section~\ref{sec:introduction}).

Second, the comparison between the PL w/ index and PL w/o index conditions underscores the importance of index finger coordination, particularly when handling heavier objects. As demonstrated by the aluminum objects (Table~\ref{tab:Exp1_Results_Combined}), the success rate for the PL w/o index decreased markedly, particularly for wider objects (e.g., 60\% for 15~mm prism and 40\% for 30~mm prism), whereas the PL w/ index maintained a considerably improved performance (e.g., $\geq$80\% for all prisms), achieving our 150~g target with a simpler design than the dexterous hands used in related work on lighter objects~\cite{Funabashi2018, Or2016}. This confirms that actively positioning the index finger to provide appropriate support against the thumb during the lateral grasp phase, which is enabled by the width estimation, substantially enhances the grasp stability and force application capabilities, overcoming a key limitation identified in our previous work~\cite{KurodaICORR2025}.

However, the experiments also revealed the limitations of our approach. The lower success rates for the cylinders compared with those of the prisms likely originates from a different grasp instability aspect, which stems from the difficulty in establishing a stable contact area on a curved surface. This highlights the trade-off in our optimization method. We evaluated grasp stability based on the conditions indicated in our previous work~\cite{KurodaICORR2025}, which assumes point contact for computational tractability. However, as noted in prior studies, grasps planned with point-contact models are vulnerable to these types of rotational moments~\cite{Bicchi2000}, a problem that is more pronounced on the point contact of a cylinder than on the surface contact of a prism~\cite{Veiga2020}. Further refinement of this optimization is another important direction for future research.

\subsection{Experiment 2: Demonstration of Practical Applications}
The practical utility of the PL manipulation capability (using the PL w/ index strategy) was demonstrated via several tasks representative of ADLs. The successful execution of these tasks is demonstrated in Fig.~\ref{fig:Demonstrations}, as well as supplementary Movies S1 (https://youtu.be/TswoBs84uyE), which highlight the potential benefits of this system.

The demonstrations in Experiment 2 qualitatively confirmed the practical potential of the manipulation system. By controlling the PL transition based on simple current feedback, the PLEXUS hand could perform a wider range of functional tasks compared with scenarios requiring manual reorientation of the object with the other hand or where static grasps are only used (as illustrated in Fig.~\ref{fig:Demonstrations}). Tasks such as turning knobs, closing caps, and orienting a marker for writing were successfully executed, suggesting its potential for reducing compensatory user movements. This success also validates our control approach, in which in-hand manipulation is autonomously executed from a single user command, thereby suggesting its compatibility with prosthetic control schemes that rely on a limited set of discrete commands.

Despite the various advantages exemplified herein, this study is also associated with limitations. As noted in the bottle cap task case the maximum applicable torque for complete tightening was inherently limited by the grasp force achievable by the actuators despite the fact that the hand was able to perform the rotation sufficiently to prevent spillage. This can be insufficient for tasks requiring very high torques. Furthermore, real-world applications would necessitate seamless integration and coordination of wrist and arm movements, which constitutes an area for future work.

\section{Conclusion}
This study developed a novel lightweight electric prosthetic hand (311 g, four motors) that enables stable PL in-hand manipulation, including heavy objects, without the use of external sensors. The core achievement of this study is the demonstration of a lightweight prosthetic hand that can achieve dexterous PL manipulation via the combination of an optimized, mechanically simple thumb and a basic motor-current-based controller. This study demonstrated a viable pathway toward prosthetic hands that combine functionality with low user burden in daily use.

\addtolength{\textheight}{-12cm} 




\section*{ACKNOWLEDGMENT}
This work was supported by JSPS KAKENHI (grant numbers JP24KJ0248 and JP25K00150).

\addtolength{\textheight}{2cm}


\bibliographystyle{IEEEtran}

\bibliography{reference_2025624}

@inproceedings{BelterDollar2011,
   author = {J. T. Belter and A. M. Dollar},
   isbn = {978-1-4244-9862-8},
   booktitle = {2011 IEEE International Conference on Rehabilitation Robotics (ICORR)},
   month = jun,
   pages = {1-7},
   publisher = {IEEE},
   title = {Performance characteristics of anthropomorphic prosthetic hands},
   year = {2011},
}

@article{Feix2014,
   author = {Thomas Feix and Ian M. Bullock and Aaron M. Dollar},
   doi = {10.1109/TOH.2014.2326871},
   issn = {1939-1412},
   issue = {3},
   journal = {IEEE Transactions on Haptics},
   month = jul,
   pages = {311-323},
   pmid = {25248214},
   publisher = {Institute of Electrical and Electronics Engineers},
   title = {Analysis of Human Grasping Behavior: Object Characteristics and Grasp Type},
   volume = {7},
   year = {2014},
}

@article{Vinet1995,
   author = {Robert Vinet and Yves Lozac and Nicolas Beaudry and Gilbert Drouin},
   isbn = {07487711},
   issn = {0748-7711},
   issue = {4},
   journal = {Department Veterans Affairs Journal of Rehabilitation Research and Development},
   pages = {316-324},
   pmid = {9603206312},
   title = {Design methodology for a multifunctional hand prosthesis},
   volume = {32},
   year = {1995},
}

@inproceedings{Or2016,
   author = {Keung Or and Mami Tomura and Alexander Schmitz and Satoshi Funabashi and Shigeki Sugano},
   isbn = {978-1-5090-3762-9},
   booktitle = {2016 IEEE/RSJ International Conference on Intelligent Robots and Systems (IROS)},
   month = oct,
   pages = {2542-2547},
   publisher = {IEEE},
   title = {Position-force combination control with passive flexibility for versatile in-hand manipulation based on posture interpolation},
   year = {2016},
}

@inproceedings{Funabashi2018,
   author = {Satoshi Funabashi and Alexander Schmitz and Takashi Sato and Sophon Somlor and Shigeki Sugano},
   isbn = {978-1-5386-7283-9},
   booktitle = {2018 IEEE-RAS 18th International Conference on Humanoid Robots (Humanoids)},
   month = nov,
   pages = {1-9},
   publisher = {IEEE},
   title = {Versatile In-Hand Manipulation of Objects with Different Sizes and Shapes Using Neural Networks},
   year = {2018},
}

@article{Cipriani2010,
   author = {Christian Cipriani and Marco Controzzi and Maria Chiara Carrozza},
   doi = {10.1017/S0263574709990750},
   issn = {02635747},
   issue = {6},
   journal = {Robotica},
   month = oct,
   pages = {919-927},
   title = {Objectives, criteria and methods for the design of the SmartHand transradial prosthesis},
   volume = {28},
   year = {2010},
}

@article{Mohammadi2020,
   author = {Alireza Mohammadi and Jim Lavranos and Hao Zhou and Rahim Mutlu and Gursel Alici and Ying Tan and Peter Choong and Denny Oetomo},
   issn = {19326203},
   issue = {5},
   journal = {PLoS ONE},
   month = may,
   pages = {e0232766},
   pmid = {32407396},
   publisher = {Public Library of Science},
   title = {A practical 3D-printed soft robotic prosthetic hand with multi-articulating capabilities},
   volume = {15},
   year = {2020},
}

@article{Cipriani2008,
   author = {Christian Cipriani and Franco Zaccone and Silvestro Micera and M. Chiara Carrozza},
   doi = {10.1109/TRO.2007.910708},
   issn = {15523098},
   issue = {1},
   journal = {IEEE Transactions on Robotics},
   month = feb,
   pages = {170-184},
   title = {On the shared control of an EMG-controlled prosthetic hand: Analysis of user-prosthesis interaction},
   volume = {24},
   year = {2008},
}

@article{Carroll2004,
   author = {Áine M Carroll and Neil Fyfe},
   issn = {1040-8800},
   issue = {2},
   journal = {Journal of Prosthetics and Orthotics},
   month = apr,
   pages = {66-68},
   title = {A Comparison of the Effect of the Aesthetics of Digital Cosmetic Prostheses on Body Image and Well-Being},
   volume = {16},
   year = {2004},
}

@misc{Ossur2023,
   author = {Össur},
   title = {i-Limb Quantum},
   url = {https://www.ossur.com/en-us/prosthetics/arms/i-limb-quantum},
   note = {Accessed: Nov. 21, 2024},
}

@article{Wang2022,
   author = {Yanchao Wang and Ye Tian and Haotian She and Yinlai Jiang and Hiroshi Yokoi and Yunhui Liu},
   doi = {10.3390/mi13020219},
   issn = {2072666X},
   issue = {2},
   journal = {Micromachines},
   month = feb,
   publisher = {MDPI},
   title = {Design of an Effective Prosthetic Hand System for Adaptive Grasping with the Control of Myoelectric Pattern Recognition Approach},
   volume = {13},
   year = {2022},
}

@misc{OttobockUS2023,
   author = {{Ottobock US}},
   title = {bebionic},
   url = {https://www.ottobockus.com/prosthetics/upper-limb-prosthetics/solution-overview/bebionic-hand/\#video-2},
   note = {Accessed: Nov. 21, 2024},
}

@article{KurodaCBS2025,
   author = {Yuki Kuroda and Yusuke Yamanoi and Hai Jiang and Yoshiko Yabuki and Yuki Inoue and Dianchun Bai and Yinlai Jiang and Jinying Zhu and Hiroshi Yokoi},
   doi = {10.34133/cbsystems.0195},
   issn = {26927632},
   journal = {Cyborg and Bionic Systems},
   publisher = {American Association for the Advancement of Science},
   title = {Toward Cyborg: Exploring Long-Term Clinical Outcomes of a Multi-Degree-of-Freedom Myoelectric Prosthetic Hand},
   volume = {6},
   year = {2025},
}

@article{OpenAI2019,
   author = {OpenAI and Ilge Akkaya and Marcin Andrychowicz and Maciek Chociej and Mateusz Litwin and Bob McGrew and Arthur Petron and Alex Paino and Matthias Plappert and Glenn Powell and Raphael Ribas and Jonas Schneider and Nikolas Tezak and Jerry Tworek and Peter Welinder and Lilian Weng and Qiming Yuan and Wojciech Zaremba and Lei Zhang},
   month = {10},
   title = {Solving Rubik's Cube with a Robot Hand},
   url = {http://arxiv.org/abs/1910.07113},
   year = {2019},
}

@article{Yin2023,
  title          = {Rotating without Seeing: Towards In-hand Dexterity through Touch },
  author         = {Yin, Zhao-Heng and Huang, Binghao and Qin, Yuzhe and Chen, Qifeng and Wang, Xiaolong},
  journal        = {Robotics: Science and Systems},
  year           = {2023},
}

@inproceedings{Handa2023,
   author = {Ankur Handa and Arthur Allshire and Viktor Makoviychuk and Aleksei Petrenko and Ritvik Singh and Jingzhou Liu and Denys Makoviichuk and Karl Van Wyk and Alexander Zhurkevich and Balakumar Sundaralingam and Gavriel State and Jean Francois Lafleche},
   isbn = {9798350323658},
   issn = {10504729},
   booktitle = {2023 IEEE International Conference on Robotics and Automation (ICRA)},
   pages = {5977-5984},
   publisher = {IEEE},
   title = {DeXtreme: Transfer of Agile In-hand Manipulation from Simulation to Reality},
   year = {2023},
}

@inproceedings{KurodaICORR2025,
  title={PLEXUS Hand: Lightweight Four-Motor Prosthetic Hand Enabling Precision–Lateral Dexterous Manipulation},
  author={Kuroda, Yuki and Takahashi, Tomoya and Beltran-Hernandez, Cristian C. and Hamaya, Masashi and Tanaka, Kazutoshi},
  booktitle={2025 IEEE International Conference on Rehabilitation Robotics (ICORR)},
  year={2025},
  organization={IEEE}
}

@INPROCEEDINGS{Bicchi2000,
  author={Bicchi, Antonio. and Kumar, Vijay.},
  booktitle={2000 IEEE International Conference on Robotics and Automation (ICRA)}, 
  title={Robotic grasping and contact: a review}, 
  year={2000},
  number={},
  publisher = {IEEE},
  pages={348-353 vol.1}}

@Article{Veiga2020,
AUTHOR = {Veiga, Filipe and Edin, Benoni and Peters, Jan},
TITLE = {Grip Stabilization through Independent Finger Tactile Feedback Control},
JOURNAL = {Sensors},
VOLUME = {20},
YEAR = {2020},
NUMBER = {6},
ARTICLE-NUMBER = {1748},
PubMedID = {32245193},
ISSN = {1424-8220}
}

@article{Shim2007,
title = "Multi-digit maximum voluntary torque production on a circular object",
author = "Shim, {Jae Kun} and Junfeng Huang and Alexander Hooke and Mark Latsh and Vladimir Zatsiorsky",
year = "2007",
month = may,
language = "English",
volume = "50",
pages = "660--675",
journal = "Ergonomics",
issn = "0014-0139",
publisher = "Taylor and Francis Ltd.",
number = "5",
}

\end{document}